\documentclass[10pt,twocolumn,letterpaper]{article}

\usepackage{wacv}
\usepackage{times}
\usepackage{epsfig}
\usepackage{graphicx}
\usepackage{amsmath}
\usepackage{amssymb}
\usepackage[ruled,vlined]{algorithm2e}
\usepackage{graphics}
\graphicspath{ {./figs/} }
\usepackage{subfigure}
\usepackage{multirow}
\usepackage{enumitem} 
\usepackage{xargs}
\usepackage{comment}
\usepackage{caption}
\usepackage{bbm}
\usepackage{pifont}
\usepackage{mathtools}
\usepackage{microtype}

\def\wacvPaperID{****} 

\wacvfinalcopy 

\ifwacvfinal
\fi

\ifwacvfinal
\usepackage[breaklinks=true,bookmarks=false]{hyperref}
\else
\usepackage[pagebackref=true,breaklinks=true,colorlinks,bookmarks=false]{hyperref}
\fi

\begin{document}

\title{Fair and accurate age prediction using distribution aware data curation and augmentation}

\author{Yushi Cao$^{1,a}$\thanks{Yushi Cao and David Berend cobntributed equally.}, David Berend$^{1,3,a}$, Palina Tolmach$^{1,4}$, Guy Amit$^2$, Moshe Levy$^2$\\
Yang Liu$^1$, Asaf Shabtai$^2$, Yuval Elovici$^2$\\
$^1$Nanyang Technological University, $^2$Ben-Gurion University of the Negev\\
$^3$ Singapore Institute of Manufacturing and Technology, A*STAR\\
$^4$Singapore Institute of High Performance Computing, A*STAR\\
{\tt\small \{yushi002, bere0003, palina0001\}@e.ntu.edu.sg, \{guy5, moshe5\}@post.bgu.ac.il}\\
{\tt\small yangliu@ntu.edu.sg, \{shabtaia, elovici\}@bgu.ac.il}
}

\maketitle
\ifwacvfinal
\thispagestyle{empty}
\fi

\begin{abstract}
Deep learning-based facial recognition systems have experienced increased media attention due to exhibiting unfair behavior. Large enterprises, such as IBM, shut down their facial recognition and age prediction systems as a consequence. Age prediction is an especially difficult application with the issue of fairness remaining an open research problem (e.g., predicting age for different ethnicity equally accurate). One of the main causes of unfair behavior in age prediction methods lies in the distribution and diversity of the training data. 
In this work, we present two novel approaches for dataset curation and data augmentation in order to increase fairness through balanced feature curation and increase diversity through distribution aware augmentation. To achieve this, we introduce out-of-distribution detection to the facial recognition domain which is used to select the data most relevant to the deep neural network's (DNN) task when balancing the data among age, ethnicity, and gender. Our approach shows promising results. Our best-trained DNN model outperformed all academic and industrial baselines in terms of fairness by up to 4.92 times and also enhanced the DNN's ability to generalize outperforming Amazon AWS and Microsoft Azure public cloud systems by 31.88\% and 10.95\%, respectively.

\end{abstract}

\begin{figure}[t]
  \centering
  \includegraphics[scale=0.7]{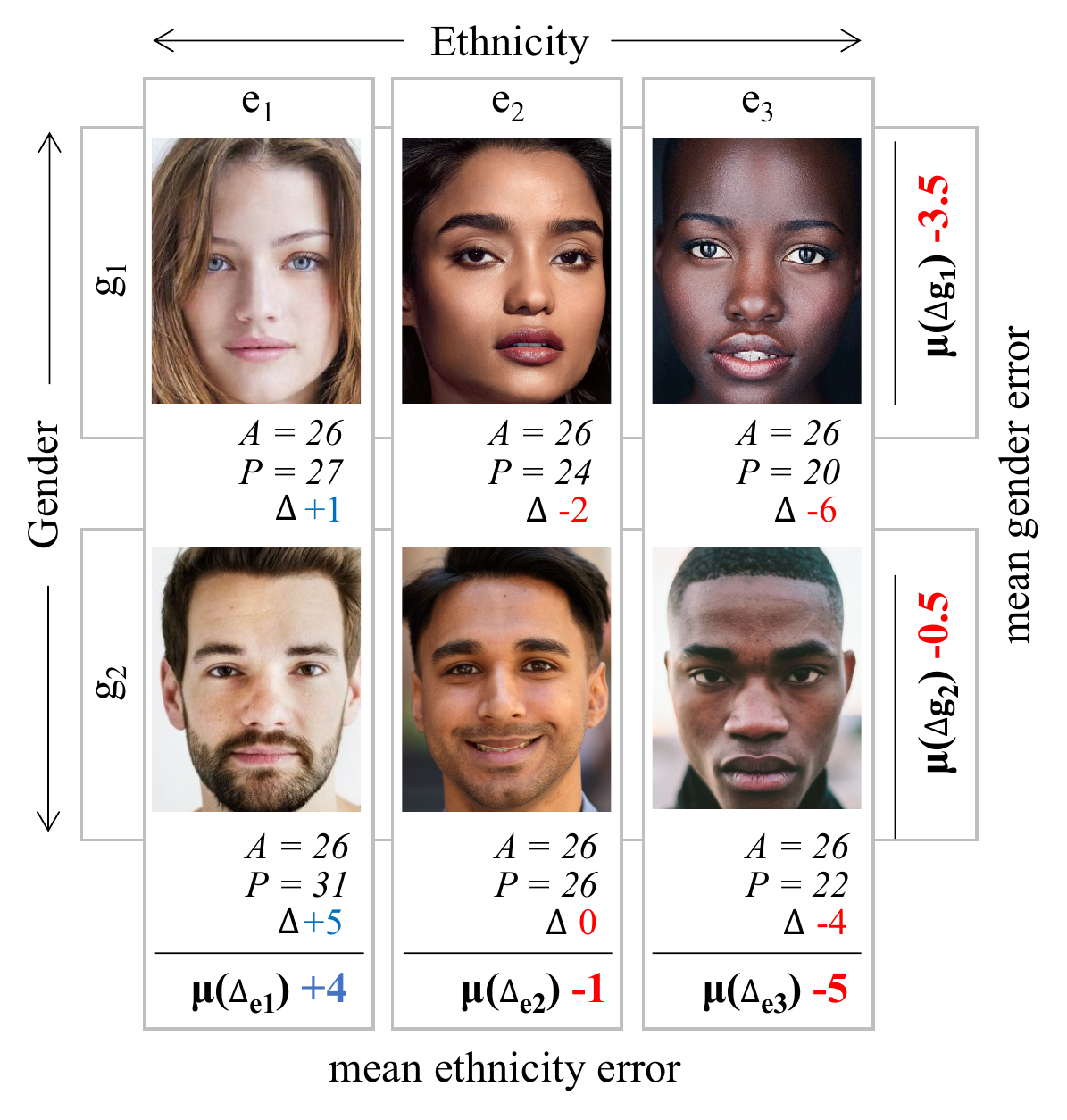}
  \caption{Comparison of actual (A) and predicted age (P). Illustrates existence of systematic unequal prediction among gender and ethnicity which can be avoided.}
  \label{fig:overview}
  \vspace{-15pt}
\end{figure}

\section{Introduction}
Due to their high accuracy, deep learning-based facial recognition systems have been widely adopted by governments and industry, with \$3.8 billion in governmental spending on such systems in 2020~\cite{face_rec_market}. Despite this government investment and their accuracy and rate of adoption, there have been numerous cases in which deep learning (DL) systems have been shown to behave unfairly. More specifically, in some cases, DL systems have demonstrated poor accuracy for a specific ethnicity or gender, e.g., male, while demonstrating high accuracy for other feature representations, e.g., female \cite{AppleIncident, amazon_customers, Facewatch}. This inconsistent accuracy has led large enterprises, such as IBM, to eliminate the use of facial recognition systems~\cite{IBM}. Amazon Web Services (AWS) and Microsoft Azure have followed this trend, limiting the use of their facial recognition systems to governments and more specifically, to law enforcement agencies~\cite{amazon_customers,Microsoft}. However, the development of facial recognition systems continues and their adoption is increasing~\cite{face_rec_market}. Therefore, there is a need to ensure the fairness of existing facial recognition systems and those developed in the future.

Age prediction is considered one of the most challenging facial recognition system tasks. The difficulty stems from the high number of predictable classes representing ages, among which there are small apparent physical differences~\cite{fairnessperf:baseline}. Age prediction systems have also been widely adopted, with applications in law enforcement~\cite{police}, surveillance~\cite{survey:agepred,survey:agepred2}, marketing~\cite{survey:agepred,market1}, and identity acquisition~\cite{police,survey:agepred}. These systems are considered unfair when the mean of the predicted age is different for, e.g., male and female~\cite{survey:bias}.


To provide an example of unfair DL age prediction systems and their devastating impact, we present a law enforcement use case. In this setting, age is a factor, as younger people are more likely to commit a crime than older people. Law enforcement agencies have increased their use of DL-based age prediction systems~\cite{LawEnforcement}. These systems may be inherently unfair and consistently predict that African Americans are  younger than Caucasians, and as a result, African Americans may be unfairly punished by law enforcement agencies solely due to their ethnicity. In fact, just recently news has reported that COMPAS, a tool used in many US jurisdictions to help make decisions about pretrial release and sentencing, was twice as likely to favor arrests for the African American ethnicity than for Caucasian ethnicity~\cite{racistpolicing,racistCOMPAS}.

Age prediction belongs to the field of computer vision (CV) where the issue of fairness is a challenging task due to the difficulty in creating diverse and sensitive feature balanced datasets. Most features and camera settings are embedded into the image as a combination of pixels, rather than labels. One simple solution is to combine all available data from existing benchmarks into one data pool. However, while this large data pool may be more diverse, it can still be imbalanced in its distribution of features and inherit the bias of an individual dataset. This is a common challenge faced by large cloud systems~\cite{watson,aif360} 
and by academia, reflected by the imbalanced distribution behavior in academic age prediction benchmark datasets used in prior studies~\cite{DEX,relwork:bias_existance}. These benchmark datasets show fundamentally different distribution of features and camera settings which makes comparison of age prediction approaches using different benchmark datasets an infeasible task.

To address the challenge of curating diverse and balanced datasets, we propose a novel dataset curation approach that increases diversity by maximizing the balance between sensitive features. In addition, we introduce distribution awareness~\cite{ood:baseline} to the age prediction domain. Distribution awareness provides a high-dimensional certainty estimate from the neural network for each input. Therefore, we aim to improve data augmentation which is commonly used to increase the amount of data to ensure that the DNN model has learned sufficient representations of sensitive features. We hypothesize that not all augmentation may be beneficial to a DL system's learning process and may in fact perpetuate existing bias in the data. This benefits our balancing efforts as we can enhance the dataset with images related to features that have insufficient samples while ensuring the added augmentation are indeed beneficial to the age prediction system.


In summary, our contributions are as follows:
\begin{itemize}
\item We propose feature-aware dataset curation, an approach that aims to increase fairness by maximizing the balance among sensitive features. 
\item We introduce distribution awareness to data augmentation to further increase diversity and sufficiency of sensitive features. We find that not all augmentations benefit performance and that these harmful augmentations are in majority far from the trained data distribution and can be identified and filtered out using our lightweight out-of-distribution detection technique. 
\item We are the first in age prediction research to evaluate our approaches on a large scale, using 24 DNN models, six benchmark datasets, and one million data points. This enables a fair comparison in which we outperform  approaches of prior research~\cite{DEX, fairnessperf:baseline} 
and those of Amazon AWS and Microsoft Azure in fairness and performance\footnote{Our code, DNN models, and public-accessible data are available in our public repository: https://github.com/ForBlindRev/AIBias}.

\end{itemize}

\section{Background \& Related Work} \label{sec:background}
\subsection{Age Prediction}

With the increase in computing capacity and the rise of deep learning, new approaches rise to advance the age prediction field~\cite{ageprec:rankCNN,background:age-pred2,background:age-pred3,ageprec:survey}. Yi et al.~\cite{ageprec:MSCNN,ageprec:CNN} was one of the first to utilize convolutional neural networks to extract facial features to estimate age. 
The best performance on the benchmark dataset MORPH-2 was achieved by Zhang et al.~\cite{ageprec:LSTM} who utilized a network called AL-RoR which combines long- and short-term memory (LSTM) architecture~\cite{rel:lstm} and residual networks~\cite{arch:resnet50} to achieve a 2.36 MAE. A study performed by Rothe et al.~\cite{DEX} presented a method based on common classification networks, such as VGG-16~\cite{arch:vgg16}, which is transferable to ResNet-50~\cite{arch:resnet50} or DenseNet-121~\cite{arch:densenet}, making it easily reproducible and which obtained state-of-the-art performance for age prediction~\cite{relwork:paperswithcode} on the MORPH dataset, with an MAE of 2.68.
We use the approach of Rothe et al. as baseline.

\begin{figure*}[t]
  \centering
  \includegraphics[width=1.7\columnwidth]{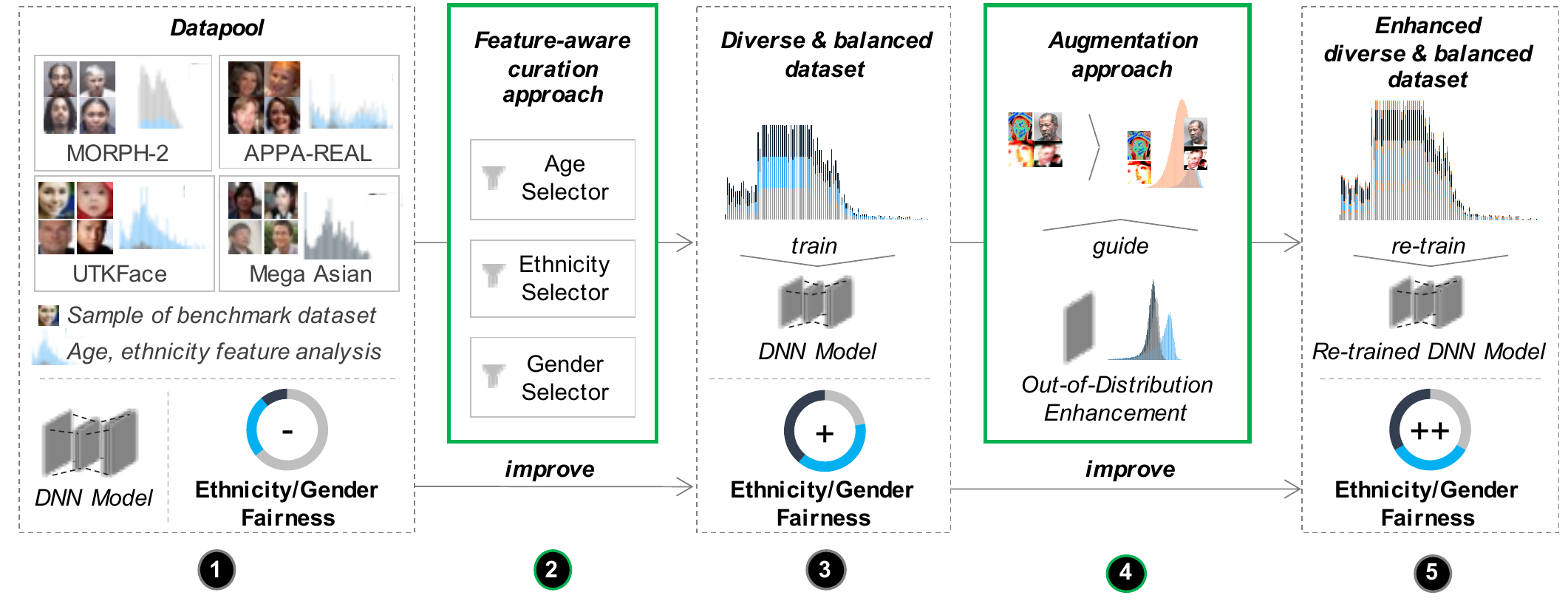}
  \caption{Overview of our approaches.}
  \label{fig:overview}
  \vspace{-10pt}
\end{figure*}

\subsection{Fairness \& Mitigation}\label{sec:background-fairness}

Addressing the fairness of age prediction systems is challenging. This may be due to the fact that sensitive features are not always available as labeled data. Instead, they are inherent in the picture and represented by pixels. 
A study performed by Clapés et al.~\cite{fairnessperf:baseline} aimed at fair age prediction; the authors also analyzed the difference between the real age of a person and their apparent age. 
Their proposed fairness improvements involve shifting the apparent age towards the actual age during training. The results of the study are somewhat questionable, as the reported performance measured in terms of the MAE varied by 12 years, on average. 
In other DL domains several mitigation techniques for addressing fairness exist~\cite{relwork:bias_existance,relwork:whitebox,relwork:L2-loss, fakespotter, spark, fakepolisher, deeprhythm, forecasting, instance_learning}. Given the additional features, e.g., DL loss functions can be optimized towards fair representations \cite{relwork:bias_existance,relwork:L2-loss}. However, such methods remain so far limited in the field of age prediction. This calls for new ways to evaluate and optimize such systems which goes beyond previous efforts focusing primarily on performance, e.g., by NIST using the FERET dataset~\cite{relwork:feret}.

\subsection{Distribution Awareness}
Distribution awareness, or also referred to as Out-Of-Distribution (OOD) detection \cite{ood:baseline} is the ability to compare what the DNN model has learned to what the DNN model receives as input. The first goal is to filter out data that is not suited for the DNN application, both during training and later during testing and deployment. The second goal is to select data that fits the DNN application and inherits new information the DNN model has not learned yet, e.g., balance ethnicity or gender by selecting diverse yet realistic mutations.
Recently, various techniques have been  proposed as OOD scoring functions based on the DNN model's activation, such as~\cite{ood:llratio, ood:llratio2, ood:odin, ood:gen_ensemble, ood:scale_ensemble, ood:reject_class, ood:conf_cal, DBLP:conf/iclr/HendrycksG17, ood:cos_sim, ood:typicality, ood:maha, ood:oe}. These techniques provide different ways of calculating a score on how far or close a given input is to the trained data distribution. Hendrycks et al.~\cite{ood:baseline, ood:oe} used the maximum softmax probability of the DNN model output. Liang et al.~\cite{ood:odin} added an additional input perturbation to retrieve the OOD score. Others~\cite{ood:reject_class, ood:llratio, ood:llratio2} utilized separately trained DNN models or changed the DNN model architecture to calculate the OOD score. 

\section{Methodology}
\subsection{Overview}
Figure \ref{fig:overview} presents the overview of our approaches. We present a novel dataset curation approach and a novel data augmentation approach (both indicated in green). Overall, the methodology consists of five steps. In step \ding{182}, we analyze the most recent age prediction benchmark datasets which tend to be unbalanced and poorly distributed; to do this, we cross-validate four benchmark datasets along with two DNN training approaches of prior research. The four benchmark datasets have varying diversity and varying quantities of sensitive features which will help in identifying the root cause of unfair behavior. In order to curate a diverse and balanced dataset, we present step \ding{183}, where we employ our novel dataset curation approach using the four benchmark datasets as a combined data pool; this simulates a data scenario commonly faced by large enterprises. 
Once the dataset is curated in \ding{184}, we further enhance the DL system in step \ding{185} where we apply our novel distribution-aware augmentation approach. Here, we aim to assess how fairness and performance can be enhanced by filtering augmentations with very high (too different from the trained distribution) and very low (too similar to the trained distribution) OOD scores. Therefore, we re-train the DNN model in step \ding{186} with the respective augmentation sets to identify optimal data distributional setting for maximizing fairness and performance. To validate the effectiveness, we evaluate our final age prediction system to those following approaches of the previous state of the art in academia and to the leading facial recognition systems from Amazon AWS and Microsoft Azure, using two individual benchmark datasets for assessing generalization.

\subsection{Feature Aware Dataset Curation}\label{curation}

We aim to curate a dataset which is both diverse and well-balanced among sensitive features. To accomplish this, we propose a dataset curation approach. Given a set containing all the data sources $\{d_{1},\ldots,d_{n}\}$, we denote the union of all data sources by $\mathcal{D}$, the set of sensitive features $S$, and $L$ the set of labels, we select the samples where $x_{i}^{s}$ is the subset of samples containing the feature $s$ with value $i$ , and $x_{i, d_{t}}^{L}$ are the samples labeled as $a$, a subscript $d_{t}$ denotes that the samples originated from data source $d_{t}$:

\begin{equation}
\begin{split}
&\mathop{\arg\max}_{x\subseteq \mathcal{D}} |x|\text{  s.t.  } \forall t,z\leq n :\\
&\forall a,b \in L : |x_{a}^{L}| = |x_{b}^{L}| \wedge |x_{a, d_{t}}^{L}| =  |x_{a, d_{z}}^{L}| \\
&\forall s\in S, \forall i,j \in s:  |x_{i}^{s}| = |x_{j}^{s}| \wedge |x_{i, d_{t}}^{s}| =  |x_{i, d_{z}}^{s}|
\end{split}
\end{equation}

 Thereby, we aim to balance data among sensitive features for each age while extracting data from as many data sources to maximise diversity. In case insufficient data exists for a given age we define a global maximum (to prevent overfitting) and minimum (to prevent bias) for selecting samples for a given age. The pseudocode of the data curation can be found in the appendix A.5.

\subsection{Distribution Aware Augmentation}

Effective dataset curation relies on data availability; the ability to address data insufficiency and increase the amount of available data and help improve the balance between sensitive features. This improves fairness, and/or improve the uniformity of available data per age, thereby improving performance. Therefore, data augmentation is commonly used to address data insufficiency when creating a DL system. However, one drawback of augmentation is that even with careful consideration of the augmentation parameters, there is no guarantee that all augmentations will be considered realistic or feasible for the trained distribution or application task of the DNN model~\cite{ood:catfish, rel:deephunter}. Furthermore, some augmentations may not modify the original image enough, producing too many similar images which may result in overfitting.
Different ways to create augmentations have been studied. We illustrate and compare the most prominent approaches in the appendix A.4. and continue with the best performing approach which employs both affine and color augmentations~\cite{ood:catfish, rel:deephunter, rel:dlfuzz}.

One of our key novelties is the utilisation of OOD scores when selecting augmentations. Here, we opted for the base methodology of the fast-out-of-distribution detection method (FOOD)~\cite{ood:glod}, which is computationally efficient, does not need to train separate approaches unlike other approaches~\cite{ood:oe, ood:llratio2, ood:gen_ensemble, ood:scale_ensemble} nor requires additional data~\cite{ood:oe, ood:llratio2,ood:conf_cal}. The main idea behind FOOD is replacing the last-fully connected layer with a Gaussian likelihood layer to represent the training data of the model as a multivariate Gaussian with a center vector and a co-variance matrix as parameters. As a result, a likelihood ratio (LLR) can be extracted for each data point which serves as OOD score. We optimize this approach, such that the center $\mu_{c}$ and the co-variance $\Sigma_{c}$ can be directly calculated without requiring additional training of the Gaussian layer. Thereby, increasing overall applicability of the technique and enabling faster integration into ML workflows (the mathematical foundations can be found in the appendix A.1, the general technique is available in our repository).

We integrate the augmentation and OOD scoring capabilities in Algorithm \ref{algo:augset}.
First, $median\_num$, $mean\_num$, and $max\_{num}$ are defined by the median, average, and maximum number of samples among classes and states (lines 1-3). 
Then, $max\_ratio$ is calculated, which serves to limit the number of augmentations with respect to the overall dataset size and the individual age sample size (line 4). Next, for each class $c$ and sensitive feature $s$, $aug\_ratio_{c,s}$ is calculated specifically to identify and generate the required number of augmentations to enhance the balance among $s$ without overfitting to a specific representation (lines 5-12). Finally, the OOD scores are calculated by which all augmentation are sorted and filtered (line 13-14). 
This enables filtering augmentations with a large OOD score, far from the trained distribution, and filtering augmentations with a low OOD score, close to the trained distribution. Thereby, we aim to evaluate under which settings, too different or too similar augmentations can be identified and how they impact fairness and performance.
Then, the augmentations are sampled following the same approach as in \ref{curation} (lines 15-20).

\begin{algorithm}[t]
\SetAlgoLined
\LinesNumbered
\KwResult{Curated dataset enhanced by distribution-aware augmentation}
 $median\_num \leftarrow$ median samples in dataset\;
 $mean\_num \leftarrow$ average samples in dataset\;
 $max\_num \leftarrow$ maximum samples in dataset\;
 $max\_ratio \leftarrow ceil(max\_num/mean\_num)$\;
 \For{\textbf{all} $c \in C$}{
    \For{\textbf{all} $s \in S$}{
     $num\leftarrow$ number of samples in $dataset_{c,s}$\;
     $aug\_ratio_{c,s} \leftarrow ceil(median\_num/num)$\;
     $aug\_ratio_{c,s} \leftarrow min(aug\_ratio_{c,s},max\_ratio)$\;
     augmenting data according to $aug\_ratio_{c,s}$\;   
    }
 }
 calculating $OOD\_scores$ from OOD\_analysis($augmented\_data$)\;
 filter $augmented\_data$ by $OOD\_scores$\;
 \For{\textbf{all} $c \in C$}{
    get($select\_size$)\;
    \For{\textbf{all} $s \in S$}{
        sample$(augmented\_data_{c,s},select\_size)$\;
    }
 }
 \caption{Enhancing a dataset by distribution-aware augmentation.}
 \label{algo:augset}
\end{algorithm}

\section{Evaluation}\label{sec:evaluation}

The evaluation follows Figure \ref{fig:overview}, in which we first cross-analyze commonly used benchmark datasets; this is followed by the dataset curation and augmentation evaluation.
Finally, the overall outcome of our approaches is evaluated with regard to fairness and performance, and compared with the state of the art in research and in industry.

\subsection{Experimental Setup}\label{sec:setup}

\noindent\textbf{Data.} We utilize the IMDB-WIKI dataset to pretrain our DNN models as done in related work~\cite{DEX}. Four benchmark datasets are utilized for cross-evaluation, namely APPA-REAL, MORPH-2, UTKFace and Mega Asian. To compare model fairness and performance to prior work in academia and industry, we employ two independent benchmark datasets (AFAD\cite{data:afad}, CACD\cite{data:cacd}). Details of the datasets can be found in the appendix A.2.


\begin{figure}[]
  \centering
  \includegraphics[width=1.0\columnwidth]{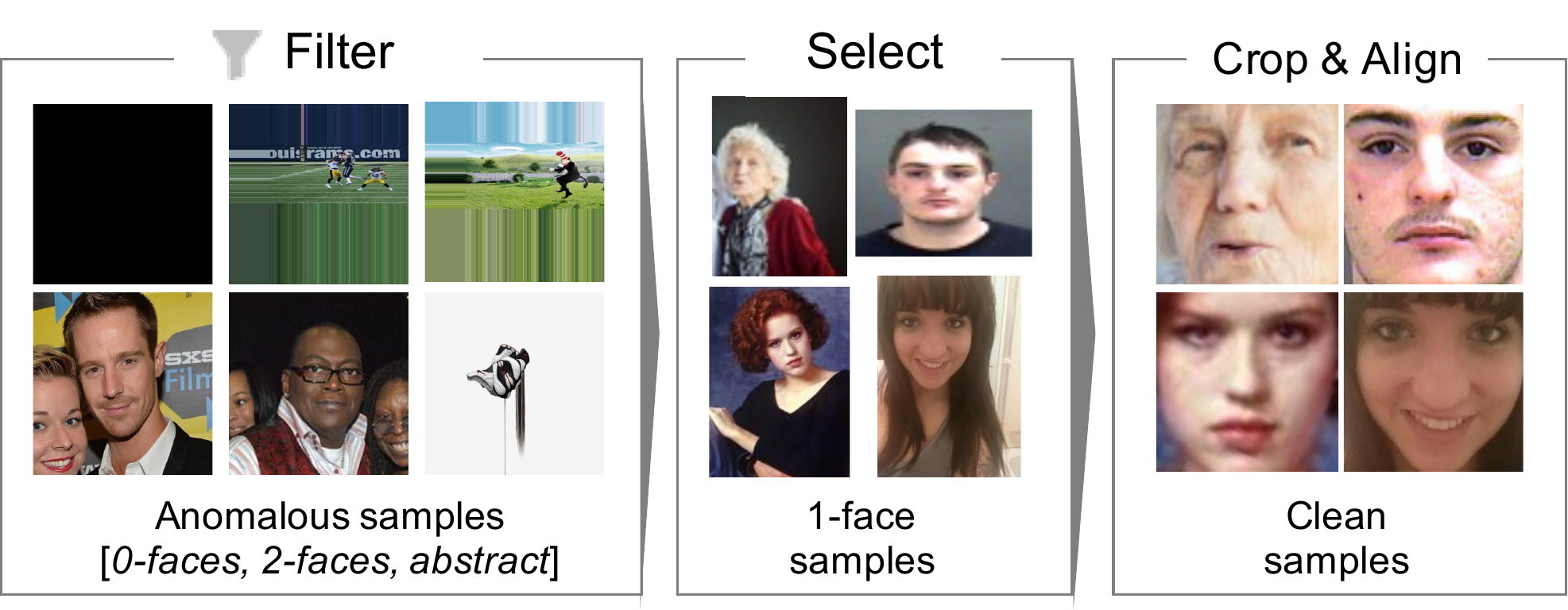}
  \caption{Pre-processing procedure.}
  \label{fig:pre_processing_examples}
  \vspace{-5pt}
\end{figure}

\noindent\textbf{Pre-processing.} For a fair comparison and to prevent dataset bias, all data follows the pre-processing procedure presented in Figure \ref{fig:pre_processing_examples}, which builds on the preprocessing procedure of ~\cite{preprocessing:dlib}.
With the help of the DNN facial recognition model, we filtered out over 10,000 of the total 758,613 images, most of which stemmed from the more noisy IMDB-WIKI benchmark dataset. Then, two common facial recognition procedures, namely crop and align, are employed on the resulting clean dataset.


\noindent\textbf{DNN Models.} In this work, a total of 24 DNN models are trained. We employ the state of the art in age prediction and computer vision for DNN model architectures. Previous research on age prediction has used the  DEX-VGG (VGG-16 based architecture)~\cite{arch:vgg16,DEX} and the AlexNet architecture~\cite{arch:alexnet}. Drawing from the general computer vision domain, we also employ the ResNet-50~\cite{arch:resnet50} and DenseNet-121~\cite{arch:densenet} DL architectures. In doing so, we aim to cover both domain-specific and general architectures to showcase the potential generalization of our approach.

\noindent\textbf{Evaluation metrics.}\label{EvalMetric}
Fairness and performance are at the core of DL systems exposed to human-centric applications like age prediction. For performance, we opt for the commonly used mean absolute error (MAE)~\cite{evaluation_mae,evaluation_mae2}, which calculates the mean of how many years the actual age has been mispredicted in absolute terms. For fairness, we build on commonly used mean distance~\cite{DEX, relwork:whitebox} which takes the mean distance between ages per sensitive features. For age prediction in particular, this distance has to be calculated for each age. Hence, we introduce a \textit{Fairness score} for further evaluation, which takes the mean distance between sensitive features and checks if the distance is lower than a pre-defined threshold $t=3$, which stems from age discrimination based on the human rights act \cite{UDHR}. Given 100 ages for evaluation, we calculate the fairness score by the per cent of ages, where the mean distance between sensitive features is below the defined threshold (a mathematical definition can be found in the appendix A.3).

\begin{figure*}
     \centering
     \includegraphics[width=\textwidth]{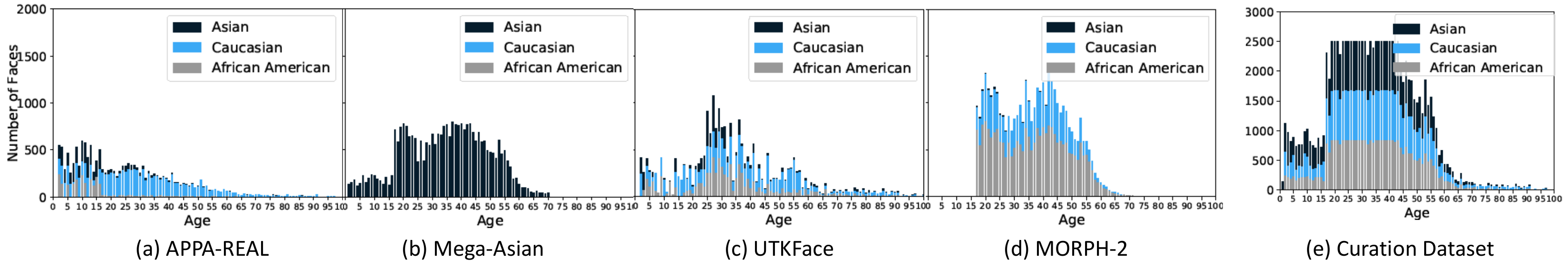}
    \caption{Sample-size per age stratified by ethnicity of the four benchmark datasets (a-d) and the curation dataset (e)}
    \label{fig:dist:dataset}
\end{figure*}

\subsection{Preliminary Cross-Analysis}\label{eva:rq1}

DL-based age prediction has relied on four commonly used benchmark datasets which differ regarding their camera setting and distribution of sensitive features. Figure \ref{fig:dist:dataset} shows the individual distributions by age segmented by ethnicity for each of the datasets. After preprocessing, MORPH-2 is the largest dataset, with predominantly faces of African American ethnicity (80\%) and a minority of Caucasian faces (18\%); the smallest group is that of Asian faces (2\%). The APPA-REAL and Mega Asian benchmark datasets are similarly imbalanced in terms of ethnicity. UTKFace is the only dataset that is somewhat balanced. 
We train two DNN model architectures for each benchmark dataset following two approaches used in prior research, namely DEX-VGG~\cite{DEX} and AlexNet~\cite{arch:alexnet}. Each DNN model is cross-evaluated with all four benchmark datasets. Table~\ref{tab:crossval} presents the average MAE of the DNN models evaluated on the training and testing data from the same benchmark dataset as well as the average of the three other benchmark datasets. We can see that the performance varies significantly depending on the benchmark dataset. For example, for MORPH-2, the average MAE across both approaches is 2.69. However, the average MAE on the other datasets is 10.10, which is the worst of the four datasets. This hints at the similar camera setting used for MORPH-2 dataset's images and the large proportion of one ethnicity in the dataset, which results in the best performance when evaluated by its equivalent testing portion but poor results when evaluated on other datasets. In contrast, when evaluating the DNN models trained on the more diverse UTKFace dataset individually, a slightly lower MAE of 5.23 is obtained. However, the average performance on the other datasets improves, decreasing the average MAE by 20.09\%. Detailed results are shown in appendix A.6.

Prior research did not compare the proposed approaches on multiple datasets. We are the first to do this and based on our comparison we observe that a fair comparison between approaches using only one but different benchmark dataset is not possible and requires diverse and balanced data.

\begin{table}[]
    \centering
    
    \caption{Cross-evaluation of prior research with benchmark datasets.}
    \label{tab:crossval}
    \resizebox{0.7\columnwidth}{!}{%
    \begin{tabular}{ccc}
    \hline\hline 
    Training Data & Testing Data & MAE$\downarrow$\tabularnewline
    \hline\hline
    \multirow{2}{*}{APPA-REAL} & APPA-REAL & 7.27 \tabularnewline
    & Others (average) & 8.74 \tabularnewline
    \hline
    \multirow{2}{*}{Mega Asian} & Mega Asian & 5.03 \tabularnewline
    & Others (average) & 9.77 \tabularnewline
    \hline
    \multirow{2}{*}{MORPH-2} & MORPH-2 &\textbf{ 2.69} \tabularnewline
    & Others (average) & 10.10 \tabularnewline
    \hline
    \multirow{2}{*}{UTKFace} & UTKFace & 5.23 \tabularnewline
    & Others (average) &\textbf{ 8.07} \tabularnewline
    \hline\hline
    \end{tabular}%
    }
    \vspace{-5pt}
\end{table}



\subsection{Feature Aware Dataset Curation}

We present a balanced and feature-aware dataset curation approach. To create our data pool, we combine the four previously mentioned benchmark datasets, whose individual data distribution by age is presented in Figure \ref{fig:dist:dataset} (a-d).
Three dataset curation baselines are used for comparison. The first baseline is no curation, called \textit{None}, for which the average results are obtained when the DNN models from Section \ref{eva:rq1} are evaluated individually. In the second baseline of random curation the total dataset size of the proposed algorithm's outcome is used to select data randomly from the pool, named \textit{Random}. The third baseline is balanced with regard to its age distribution but selects samples without focusing on balancing sensitive features, named \textit{Age only}. Our approach, named \textit{Ours}, balances both age and sensitive features. The final data distribution by age is illustrated in Figure \ref{fig:dist:dataset} (e).

The results presented in Table \ref{tab:curation} show that the dataset curated using our approach performs best in all categories (row 4, \textit{Ours} setting). This shows the effectiveness when balancing age and sensitive features from all dataset sources, diversity is maximally  increased; this benefits MAE to be well maintained across different camera setups. More importantly, the highest levels of fairness are achieved, with a fairness score of 67.50 for ethnicity and 81.00 for gender. Compared to the baseline setting \textit{None}, this represents a 188.71\% improvement in fairness for ethnicity and a 32.24\% improvement in fairness for gender.

\begin{table}[]
    \centering
    \caption{Evaluation of dataset curation setting (Ethnicity, Gender in Fairness-Score).}
    \label{tab:curation}
    \resizebox{0.8\columnwidth}{!}{%
    \begin{tabular}{ccccc}
    \hline \hline 
    Curation Setting & MAE$\downarrow$  & Ethnicity$\uparrow$ & Gender$\uparrow$\tabularnewline
    \hline \hline 
    None & 5.05  & 23.38 & 61.25\tabularnewline
    Random & 4.03  & 50.25 & 76.50\tabularnewline
    Age only & 4.06  & 63.50 & 78.50\tabularnewline
    Ours & \textbf{4.00} & \textbf{67.50} & \textbf{81.00}\tabularnewline
    \hline \hline 
    \end{tabular}%
    }
\end{table}

\subsection{\textbf{Distribution Aware Augmentation}}\label{sec:eva:aug}

\begin{figure}[t]
  \centering
  \includegraphics[scale=0.4]{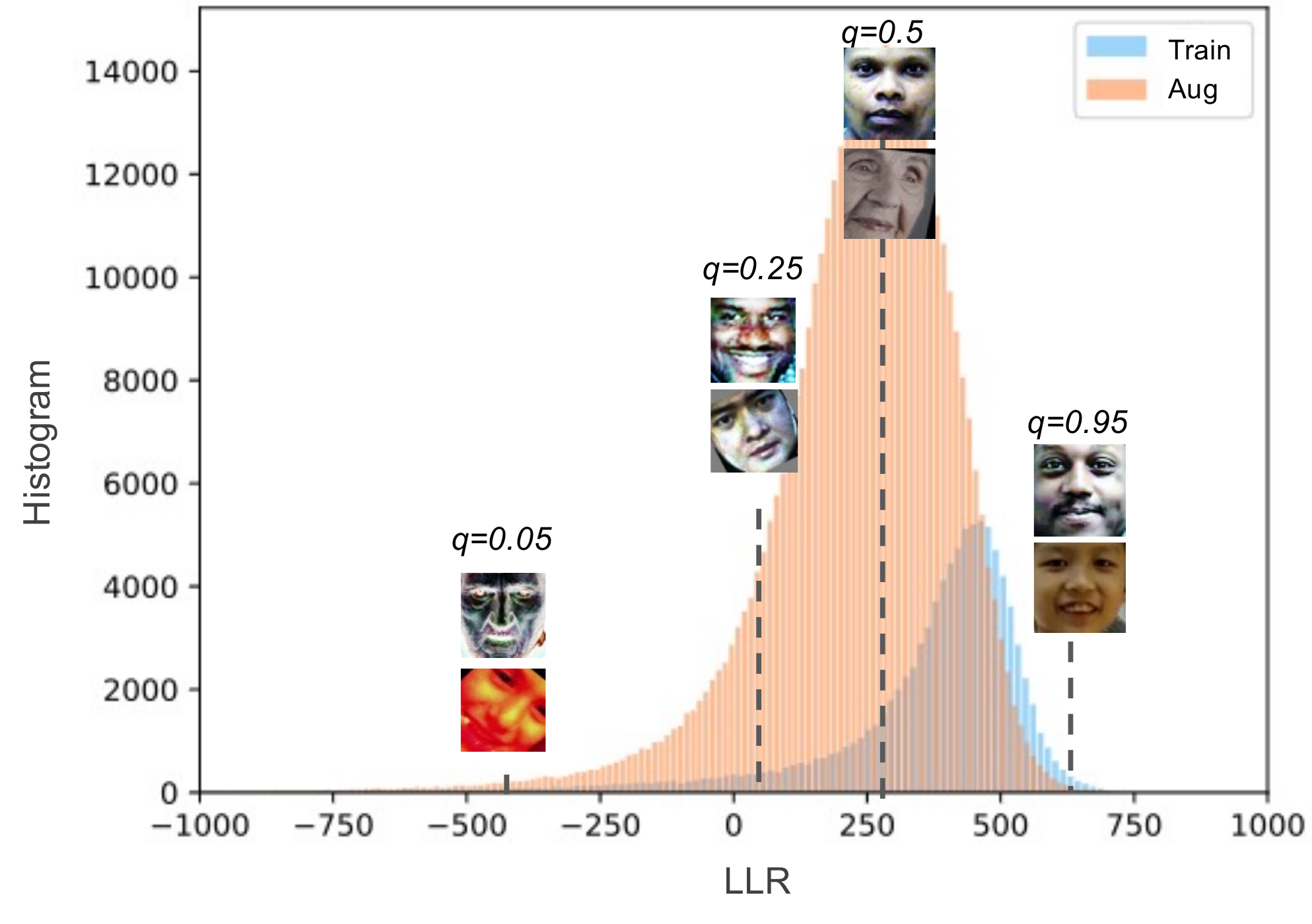}
  \caption{OOD-score distributions of augmentation data (Aug) and training data (Train) with augmentation examples based on LLR quantile.}
  \label{fig:dist:aug}
  \vspace{-5pt}
\end{figure}

Following the dataset curation we are able to increase performance above presented baselines, however, Figure \ref{fig:dist:dataset} (e) shows various insufficiencies of available data per age. 
Hence, to address the remaining data points, we augment the data (using fine-grained affine and color mutations based on~\cite{ood:catfish, rel:deephunter, rel:dlfuzz}) and add it to the dataset, which is a common practice in DL system development. We hypothesize that augmentations with a high OOD score may harm the overall DNN model's performance. A high OOD score may indicate an augmented image far from trained distribution and may be considered unrealistic in real-world scenarios. 
Figure \ref{fig:dist:aug} demonstrates this phenomenon, where similar and very different augmentations can be identified based on their OOD score. Those augmentations with a likelihood-ratio (LLR, FOOD's OOD score metric) below the 0.05 quantile (indicating a high OOD score) differ greatly from the original samples. On the other hand, those with a large LLR above 0.95 quantile (a low OOD score) have great similarity to the original samples. This is also reflected by the overlap of the training data (Train) and augmentation data (Aug) distributions. Thus, the results are in line with qualitative human-based perspective.

To quantify the impact of distribution aware augmentation, we employ two independent datasets (CACD \cite{data:cacd} and AFAD \cite{data:afad}). In this way, we aim to conduct a fair comparison. Furthermore, utilising the two datasets is close to simulating a real-world scenario in which unknown scenes with all kinds of camera settings and sensitive features are encountered, which helps assessing the DNN model's ability to generalize. The goal is to assess for which distribution aware filter setting, the best  fairness and performance is achieved. Table \ref{tab:aug:filter} presents the results for five filter settings. For all settings in which mutations are used, we generate 100,000 mutations and select 20,000 of them based on the strategy of the setting. Then, we add them to the 107,404 image large curated dataset. Afterward, the corresponding DNN models are retrained. 
The first setting, \textbf{1}, serves as comparison when no augmentations are utilised. Here, the DNN models are trained using data from the previous data curation evaluation with the  107,404 images only (first baseline). In the second setting, \textbf{2}, the mutations are randomly sampled without awareness of the data distribution, meaning data of LLR quantiles 0.00 - 1.00 are excepted (second baseline). Settings \textbf{3-5} are used to assess our approach when integrating distribution-awareness into the data augmentation process. Here, we filter all augmentations based on their OOD score first, before selecting 20,000 of them for retraining. Setting \textbf{3}, filters augmentations far from the trained distribution, which tend to look different, meaning data of LLR quantiles 0.00 - 0.05 are filtered out. Setting \textbf{4}, filters augmentations far and close to trained distribution, meaning images which tend to look different (LLR quantiles 0.00 - 0.05) and very similar (LLR quantiles 0.95 - 1.00). Finally, setting \textbf{5} serves as contrast to setting \textbf{4}, filtering all augmentations except those which are far and close to the trained distribution (LLR quantiles 0.05 - 0.95). Thereby, the goal is to analyze their potentially harmful impact.
In the table, column 2 contains the quantile range of the LLR used for augmentation selection. Columns 3-4 contain the MAE for the CACD and AFAD datasets respectively, and columns 5-6 contain the fairness score for ethnicity and gender respectively.

The results show that the best setting is \textbf{3}, when augmented data far from the trained distribution is filtered. Compared to the first and second baseline, for the CACD, the MAE improved by decreasing from 4.77 for setting \textbf{1} (no augmentation) and 4.65 for setting \textbf{2} (no filter) to 4.53 (setting \textbf{3}, filtering very different data). Similar behavior is observed for the AFAD benchmark dataset. One further observation when comparing setting \textbf{3} to setting \textbf{1} is that the fairness score for ethnicity increases from 70.50 to 73.50 and is nearly retained for gender (81.00 vs. 80.00). Filtering out very different and similar data (setting \textbf{4}) tends to under perform in comparison to filtering very different data only. While there is a risk of overfitting with similar data, the DNN model seems to benefit from the increase in data availability more at this point. Finally, setting \textbf{5} validates our results, showing that integrating very different and similar augmentations only harms DNN model fairness and underperforms in comparison to all other filter settings.



\begin{table}[t]
    \centering
    \caption{Evaluation of distribution aware augmentation approach under different OOD-score filter settings defined by LLR ranges (CACD, AFAD in MAE, Ethnicity, Gender in Fairness-Score).}
    \label{tab:aug:filter}
    \resizebox{1\columnwidth}{!}{%
    \begin{tabular}{cccccc}
    \hline \hline 
    Setting & Trained LLR Range & CACD$\downarrow$ & AFAD$\downarrow$ & Ethnicity$\uparrow$ & Gender$\uparrow$ \tabularnewline
    \hline \hline 
    \textbf{1} & No Augmentation  & 4.77 & 7.11 & 70.50 & \textbf{81.00}\tabularnewline
    \textbf{2} & {[}0.00 - 1.00{]}  & 4.65 & 7.08 & 69.00 & 75.00\tabularnewline
    \textbf{3} & {[}0.05 - 1.00{]} & \textbf{4.53} & \textbf{7.01} & \textbf{73.50} & 80.00\tabularnewline
    \textbf{4} & {[}0.05 - 0.95{]} & 4.64 & 7.05 & 69.50 & 78.00\tabularnewline
    \textbf{5}  & {[}0.00-0.05;0.95-1.00{]} & 4.62 & 7.03 & 67.00 & 80.00\tabularnewline
    \hline \hline 
    \end{tabular}%
    }
    \vspace{-10pt}
\end{table}

\subsection{Comparison to Related Work}
The curation and the augmentation approach of this work are both evaluated on four different DNN model architectures. Having assessed the overall results, we take the best performing DNN model, namely DenseNet-121, and compare it with the state-of-the-art in academia, DEX-VGG and in industrial sector, AWS and Microsoft Azure.
To retrieve the results for industrial sector, we created accounts for Microsoft Azure and Amazon AWS gave the test data as input to their age prediction system, which gave the predicted age as output. Microsoft Azure provides a discrete age estimate, while Amazon AWS provides an age interval for which we take the median age. Table \ref{tab:final:comp} presents the results and shows that our age prediction DL system is superior, especially in terms of fairness, increasing ethnicity- and gender-based fairness respectively by 4.92 and 1.88 times when compared to AWS. Also in terms of performance, our DL system performed the best, obtaining the highest MAE with the DenseNet-121 DNN model (an MAE of 4.13 on CACD and an MAE of 6.75 on AFAD). The academic baseline DEX-VGG performed worst, which however, showed slightly better fairness scores when compared to industrial sector. For the evaluated industrial systems, Microsoft Azure's system was better in both, fairness and performance when compared to Amazon AWS' system. It achieved an MAE of 4.17, which is very similar to the MAE obtained by our DL system on CACD. However, it underperformed when compared to our DL system on the AFAD benchmark.

Our results demonstrate the effectiveness of balanced dataset curation and distribution aware data augmentation, and demonstrate the importance of feature diverse datasets. 

\begin{table}[t]
    \centering
    \caption{Evaluation of our best DL system, DEX-VGG, and Microsoft Azure and Amazon AWS (CACD, AFAD in MAE, Ethnicity, Gender in Fairness-Score).}
    \label{tab:final:comp}
    \resizebox{1.0\columnwidth}{!}{%
    \begin{tabular}{cccccc}
    \hline \hline 
    DL System & CACD$\downarrow$ & AFAD$\downarrow$ & Ethnicity$\uparrow$ & Gender$\uparrow$\tabularnewline
    \hline \hline 
    Ours & \textbf{4.13}& \textbf{6.75} & \textbf{77.00} & \textbf{88.00}\tabularnewline
    DEX-VGG~\cite{DEX} & 6.24 & 9.70 & 24.00 & 57.00\tabularnewline
    Microsoft Azure  & 4.17 & 7.58 & 34.00 & 56.80\tabularnewline
    Amazon AWS & 5.57 & 9.91 & 13.00 & 30.50\tabularnewline
    \hline \hline 
    \end{tabular}%
    }
\end{table}

\subsection{Threat to Validity}
Dataset bias may be encountered when curating a new dataset out of four benchmark datasets. To address this, we individually assessed in their distribution to ensure diversity of sensitive features, camera settings, and age and trained a seperate classifier which was unable to destinct between the datasets. For evaluation purposes, we further introduced to two independent age prediction benchmark datasets, namely the CACD and AFAD, to mitigate potential sources for dataset bias.

Using only one DNN model may also present a threat to validity. To ensure confident measures, we trained four different DNN models for each experiment, and the average across these models is presented in Tables~\ref{tab:crossval} - \ref{tab:final:comp}. 

The ground-truth for determining what data is in fact in-distribution and what data is out-of-distribution is unobtainable. When evaluation the augmentations and OOD techniques we presented a statistical evaluation and empirically assessed the visualisation of the OOD score distribution including 100,000+ augmentations respective to their distribution in Figure \ref{fig:dist:aug}.

Finally, the non-public training sets used by public cloud age prediction systems may inherit samples used in our evaluation which would dramatically increase their fairness and performance. However, we found that despite this possibility, the performance and fairness of the systems evaluated were low compared to our approach.  

\section{Discussion and Future Directions}
\noindent\textbf{Proposing age prediction techniques.} As our first step, we assessed prior research and observed that single dataset are used to evaluate prior research. In our assessment, we found that there is an imbalance of age and sensitive features in these datasets. Given this, existing approaches for age prediction are difficult to compare using a single benchmark.

\noindent\textit{Future direction:} When developing new approaches, multiple benchmark datasets should be used for comparisons. In addition, different DNN architectures should be trained for robust estimation.

\noindent\textbf{Balanced benchmarks.} We note that the great diversity of camera settings and sensitive features found in the real world has not been well addressed by existing benchmarks. 

\noindent\textit{Future direction:} Our dataset curation approach has been shown effective and could serve as a valuable resource to industry and academia, both of which often have access to large sets of samples. To support ongoing development of age prediction methods, we encourage the development of large-scale and diverse benchmark datasets. Such benchmarks can be found in other domains (e.g., ImageNet~\cite{data:imagenet} is used in the computer vision domain for general classification tasks).



\noindent\textbf{Integrating distribution awareness.} One of the main contributions of this work is to introduce distribution awareness to data augmentation and thereby improve fairness, with regard to sensitive features. 

\noindent\textit{Future direction:} To further facilitate distribution-aware DL system development and contribute to future research and improvements in the field, we have made our code publicly available~\cite{repo}. In particular, there is room for improvement in the data augmentation generation step where distribution awareness could be directly integrated in guiding the mutation criteria.

\section{Conclusion}
In this paper we presented feature aware dataset curation and introduce distribution awareness to data augmentation. Our approaches aim to increase diversity and maximize balance among sensitive features, such as ethnicity and gender. We compared our novel dataset curation approach, to three baselines which our approach all outperformed. For data augmentation, we analyzed three different augmentation techniques and assessed the different and similar augmentations individually. We found that our modified lightweight OOD-technique is indeed helpful to identify augmentations which are harmful to DNN model fairness and performance, namely those augmentations which are far from the trained distribution. The largest contribution could be made for fairness. Here, our final DL age prediction system outperformed the state of the art system from Amazon AWS or Microsoft Azure by predicting age among individual ethnicity up to 4.92 times more accurate. In addition, our DL age prediction system yielded an MAE of 4.13 on CACD benchmark dataset and 6.75 on AFAD benchmark dataset, thereby outperforming prior research and outperforming DL age prediction systems of Amazon AWS and Microsoft Azure. Overall, our work stressed the importance of curating datasets that consider the sensitive features and data diversity criteria. We gave research guidance along with future directions and provide the technique and code needed to assess both criteria to encourage further work in this important field.

\section{Acknowledgments}
We thank Professor Lei Ma from University of Alberta, Canada, Prof Zhang Tianwei and Dr. Xiaofei Xie from Nanyang Technological University, Singapore, for the helpful discussions. We also thank Professor Xavier Bresson from National University Singapore for his guidance. This research is partially supported by the European Union’s Horizon 2020 research and innovation programme under grant agreement no. 830927, the National Research Foundation, Singapore under its the AI Singapore Programme (AISG2-RP-2020-019), the National Research Foundation, Prime Ministers Office, Singapore under its National Cybersecurity R\&D Program (Award No. NRF2018NCR-NCR005-0001),NRF Investigatorship NRFI06-2020-0022-0001,  the National Research Foundation through its National Satellite of Excellence in Trustworthy Software Systems (NSOE-TSS) project under the National Cybersecurity R\&D (NCR) Grant award no. NRF2018NCR-NSOE003-0001. We gratefully acknowledge the support of NVIDIA AI Tech Center (NVAITC) to our research.

{\small
\bibliographystyle{ieee_fullname}
\bibliography{egbib}
}

\newpage
\appendix
\section{Appendix}
\setcounter{table}{0}
\renewcommand{\thetable}{\Alph{section}\arabic{table}}

\subsection{Calculation of OOD Score}\label{appendix:OODcalculation}
To extract an OOD score, FOOD creates a copy of a trained DNN model and replace the last fully-connected layer with a Gaussian likelihood layer. Usually, the DNN model is trained for a few more iterations to optimize the weights of the final layer~\cite{ood:glod}. To make it more lightweight and enable its integration, we adjust the technique such that it can be integrated in any workflow of an age prediction system without requiring additional training.

The final Gaussian likelihood layer receives the output of the penultimate DNN model layer as input. The penultimate layer is commonly used for analysis, as it contains the most processed information without limiting the feature space.
With the help of the Gaussian layer, the data is represented as a multivariate Gaussian with two parameters: a center vector and a co-variance matrix.
Given our adjustment, those two parameters can be directly calculated based on the training data for each class. For the class $c$ and penultimate representations of the dataset $X$, we calculate the center $\mu_{c}$ and the co-variance $\Sigma_{c}$ as follows:
\begin{align}
& \mu_{c} =\frac{1}{|c|}\sum_{x_{i}\in c} x_{i}  \\
& \Sigma_{c} = \frac{1}{|c|}\sum_{x_{i}\in c}(x_{i}-\mu_{c})(x_{i}-\mu_{c})^{T} 
\end{align}

\noindent with the $d$-dimensional penultimate representation, where $\mathcal{N}$ stands for the multivariate Gaussian distribution, as shown in Equation \ref{eq:gauss}.

\begin{equation}
  \begin{multlined}
    f(x| \Sigma_{c}; \mu_{c}) = \log\big(\mathcal{N}(x| \mu_{c};\Sigma_{c})  \big) = \\
    -\frac{d}{2}log(2\pi) -\frac{1}{2}log(|\Sigma_{c}|)-\frac{1}{2}(x-\mu_{c})^T\Sigma_{c}^{-1}(x-\mu_{c})
\end{multlined}
\label{eq:gauss}
\end{equation}

The closer a sample is to the class center, the higher the confidence that the input belongs to a certain class and to the trained distribution.

We calculate the OOD scores using a log-likelihood ratio ($\mathcal{LLR}$) test on the subtraction of two log-likelihood scores (Equation~\ref{eqn:llr}). The test takes the probability ratio between the log of the predicted class and the logs of the unpredicted classes, where $K$ represents the group of the $k$ class indices which do not belong to the ground truth and have the top likelihood scores $\hat{y}$.

\begin{equation}
\label{eqn:llr}
    \mathcal{LLR} = \max_{c\in \{1,\ldots, C\}}f(x|\mu_{c};\Sigma_{c}) - \frac{1}{k}\sum_{i\in K}f(x|\mu_{k};\Sigma_{k})
\end{equation}

The test provides an estimate that measures how far away the sample is from its predicted class in the penultimate representation. Samples that are too far away from their predicted class relative to other classes are given a low LLR, which translates into a high OOD score.

\subsection{Generalization Datasets}\label{appendix:data}
Neither the CACD \cite{data:cacd} or the AFAD dataset\cite{data:afad} are integral to the training or testing set; both show difference in style to the training and testing set and are collected by different sources. The CACD contains $163,446$ facial images of $2,000$ celebrities; for the AFAD, we opt to use the light version which contains $60,000$ facial images collected from various Internet sources. Table~\ref{tab:datasets} shows the summary of all the datasets we used.
The preprocessing workflow was applied to both datasets before testing (Section 4.1). 
\begin{table}[!htbp]
    \caption{Summary of datasets.}
    \label{tab:datasets}
    \resizebox{1.0\columnwidth}{!}{%
    \begin{tabular}{ccccc}
    \hline \hline 
    Name & Purpose & Size & Related Work \tabularnewline
    \hline \hline 
    IMDB-WIKI & Pretraining & 636,022 & \cite{DEX,ageprec:c3ae,ageprec:LSTM} \tabularnewline
    MORPH-2 & Curation\&Augmentation & 55,000 & \cite{ageprec:c3ae,ageprec:LSTM,ageprec:ssr,ageprec:age} \tabularnewline
    APPA-REAL  & Curation\&Augmentation & 7,591 & \cite{data:appareal,fairnessperf:baseline,ageprec:age} \tabularnewline
    UTKFace & Curation\&Augmentation & 20,000 & \cite{data:utkface,relwork:bias_existance} \tabularnewline
    Mega Asian & Curation\&Augmentation & 40,000 & \cite{data:megaasian,ageprec:ssr} \tabularnewline
    AFAD & Validation & 164,432 & \cite{data:afad,ageprec:rankCNN} \tabularnewline
    CACD & Validation & 163,446 & \cite{data:cacd,background:age-pred2,DEX} \tabularnewline
    \hline \hline 
    \end{tabular}
    }
\end{table}

\subsection{Calculation of Fairness Score}\label{appendix:fairnessscore}
Function $K$ (Equation \ref{f3}) is an indicator function that indicates fairness for one sensitive feature pair $s_j$ and $s_k$ ($k\neq j$) when the average predicted ages $P(s_j|y_i)$ and $P(s_k|y_i)$ are close enough to each other defined by threshold $t$ divided by 2 given the absolute value. $P(s_j|y_i)$ represents the average predicted age at sensitive feature $s_j$, given actual age $y_i$. Function $F$ is another indicator function that indicates for age $y_j$, if the distance of average predicted age of every pair of sensitive features are close enough to each other. Therefore, $F$ represents the overall distribution of how often the DL system performs fairly one age. Finally, $p$ summarised all ages by taking the ratio of those ages which were considered fair by $F$ and all ages together.
\vspace{-5pt}
\begin{equation}\label{f1}
p=\frac{1}{n}\sum\limits _{i=1}^{n}F(y_{i}|\mathbf{s})
\end{equation}
\vspace{-5pt}
\begin{equation}\label{f2}
F(y_{i}|\mathbf{s})=\mathbbm{1}\left(\left(\sum\limits _{j\neq k}K(s_{k},s_{j}|y_i)\right)=C_m^2\right)
\end{equation}
\vspace{-5pt}
\begin{equation}\label{f3}
K(s_{k},s_{j}|y_i)=\mathbbm{1}\left(\lvert P(s_{k}|y_{i})-P(s_{j}|y_{i})\rvert<\frac{t}{2}\right)
\end{equation}

\subsection{Comparison of Augmentation Approaches}\label{appendix:augcomp}
\begin{figure}[!htbp]
  \centering
  \includegraphics[width=1.0\columnwidth]{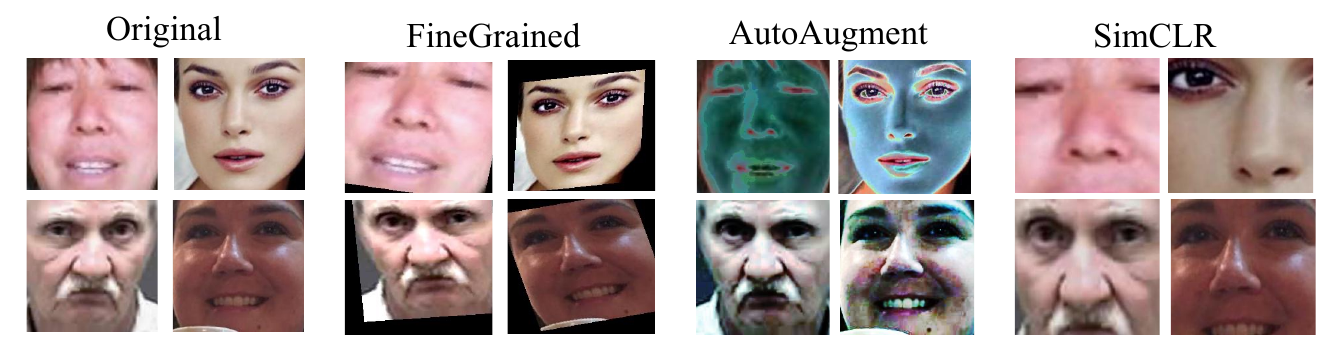}
  \caption{Example augmentations depending on augmentation type.}
  \label{fig:diffAug}
  \vspace{-10pt}
\end{figure}

In prior research, data augmentation has been assessed by identifying which augmentation types produce sufficiently diverse data for a DL system. \cite{aug:autoaugument} studied different sets of augmentation combinations to maximize diversity~\cite{aug:autoaugument}, named \textit{AutoAugment}, which is used in various prior research~\cite{aug:use_auto_1,aug:use_auto_2,aug:use_auto_3}. In the field of contrastive learning, it was found that some augmentations are beneficial when combined while others are not~\cite{aug:simclr}. As a result, the authors propose an augmentation setting used in contrastive learning to minimize the distance among augmentations from the same images while maximizing the distance among different images to determine best augmentation practices. This augmentation type is named \textit{SimCLR}.
The third augmentation type utilizes both affine and color augmentations and follows prior research~\cite{ood:catfish, rel:deephunter, rel:dlfuzz} by empirically assessing the boundaries of individual augmentations to control the diversity and realism, named \textit{Fine-grained}. Figure~\ref{fig:diffAug} shows the differences among different augmentation types. Table \ref{tab:aug:type} shows that Fine-grained method performs the best among all the settings and we mainly opt for this augmentation techniques. 

\begin{table}[!htbp]
\centering
\caption{Augmentation results comparing no augmentation setting to the presented augmentation types in Section 4.4.}
\label{tab:aug:type}
\resizebox{1.0\columnwidth}{!}{%
\begin{tabular}{ccccc}
\hline \hline 
Type  & CACD$\downarrow$ & AFAD$\downarrow$ & Ethnicity$\uparrow$ & Gender$\uparrow$\tabularnewline
\hline \hline 
None  & 4.77 & 7.11 & 70.50 & \textbf{81.00}\tabularnewline
Fine-grained  & \textbf{4.53 }& \textbf{7.01} & \textbf{73.50} & 80.00\tabularnewline
AutoAugment & 5.04 & 7.30 & 69.50 & 74.00\tabularnewline
SimCLR  & 4.58 & 7.04 & 72.00 & 69.00\tabularnewline
\hline \hline 
\end{tabular}
}
\end{table}

\newpage
\subsection{The pesudocode of data curation}\label{appendix:balancingalgo}
\begin{algorithm}[!htbp]
\SetAlgoLined
\LinesNumbered
\KwResult{Curated dataset}
$num\_sample \leftarrow $Sum of number of samples from all datasets by class $C$ and state $S$\;
sort($num\_sample$ by $s$)\;
$max\_sample \leftarrow min\{quantile(num\_sample_{c,s},0.8)|s \in S\} $\;
$min\_sample \leftarrow max\{quantile(num\_sample_{c,s},0.2)|s \in S\} $\;
\For {\textbf{all} $ c\in C$}{
    $threshold \leftarrow min\{num\_sample_{c,s}|s \in S\}$\;
    $threshold \leftarrow min(max\_sample,max(min\_sample,threshold))$\;
    $ds\_num \leftarrow$ number of datasets\;
    $select\_size \leftarrow threshold / ds\_num$\;
    \For{\textbf{all} $s \in S$}{
        sort(D,c,s)\;
        \For{\textbf{all} $d \in D$}{
            $num \leftarrow $length of $d_{c,s}$\;
            \eIf{$num < select\_size$ }{
                select all data in $d_{c,s}$\;
                $remain \leftarrow select\_size -  num$\;
                update$(select\_size,remain)$ \;
            }{
                random\_select$(d_{c,s},select\_size)$\;
            }
        }
    }
}
 \caption{Curating a diverse and sensitive feature balanced dataset}
 \label{algo:dataset}
\end{algorithm}

\newpage
\subsection{Cross-analysis results}\label{appendix:crossanalysis}
\begin{table}[hb!]
\caption{Individual cross-analysis results retrieved on prior research DL age prediction system approaches (Section 4.2).}
\label{tab:ind:ca}
\resizebox{0.9\columnwidth}{!}{%

\begin{tabular}{cclrr}
\hline \hline
\multicolumn{1}{l}{DNN} & \multicolumn{1}{l}{Train} & Test             & \multicolumn{1}{l}{MAE$\downarrow$}  \\ \hline \hline
\multirow{21}{*}{AlexNet}        & \multirow{5}{*}{APPA}              & APPA                      & 7.6                                                        \\ 
                                 &                                    & Megagsian                 & 11.8                                                         \\ 
                                 &                                    & MORPH                     & 6.5                                                          \\ 
                                 &                                    & UTKFace                   & 7.7                                                         \\  
                                 &                                    & \textbf{Average (others)} & \textbf{8.6}                                        \\  
                                 & \multirow{5}{*}{Megagsian}         & Megagsian                 & 3.6                                                           \\ 
                                 &                                    & APPA                      & 11.5                                                          \\ 
                                 &                                    & MORPH                     & 8.3                                                           \\  
                                 &                                    & UTKFace                   & 9.4                                                           \\  
                                 &                                    & \textbf{Average (others)} & \textbf{9.7}                                       \\  
                                 & \multirow{5}{*}{MORPH}             & MORPH                     & 2.9                                                       \\  
                                 &                                    & APPA                      & 11.7                                          \\  
                                 &                                    & Megagsian                 & 9.4                                                       \\  
                                 &                                    & UTKFace                   & 10.6                                                       \\  
                                 &                                    & \textbf{Average (others)} & \textbf{10.5}                                       \\  
                                 & \multirow{5}{*}{UTKFace}           & UTKFace                   & 5.3                                                        \\ 
                                 &                                    & APPA                      & 9.6                                                        \\  
                                 &                                    & Megagsian                 & 8.3                                                       \\  
                                 &                                    & MORPH                     & 7.9                                                          \\  
                                 &                                    & \textbf{Average (others)} & \textbf{8.6}                                     \\  \hline \hline
\multirow{20}{*}{DEX VGG}        & \multirow{5}{*}{APPA}              & APPA                      & 7.0                                                          \\ 
                                 &                                    & Megagsian                 & 12.2                                                         \\ 
                                 &                                    & MORPH                     & 6.4                                                          \\  
                                 &                                    & UTKFace                   & 7.9                                                          \\ 
                                 &                                    & \textbf{Average (others)} & \textbf{8.8}                                        \\ 
                                 & \multirow{5}{*}{Megagsian}         & Megagsian                 & 6.5                                                          \\ 
                                 &                                    & APPA                      & 11.4                                                 \\ 
                                 &                                    & MORPH                     & 7.3                                                           \\ 
                                 &                                    & UTKFace                   & 10.7                                                          \\ 
                                 &                                    & \textbf{Average (others)} & \textbf{9.8}                                        \\  
                                 & \multirow{5}{*}{MORPH}             & MORPH                     & 2.5                                                          \\ 
                                 &                                    & APPA                      & 10.6                                                          \\ 
                                 &                                    & Megagsian                 & 8.4                                                          \\ 
                                 &                                    & UTKFace                   & 10.0                                                     \\  
                                 &                                    & \textbf{Average (others)} & \textbf{9.7}                                         \\  
                                 & \multirow{5}{*}{UTKFace}           & UTKFace                   & 5.2                                                          \\  
                                 &                                    & APPA                      & 8.4                                                           \\  
                                 &                                    & Megagsian                 & 7.8                                                         \\  
                                 &                                    & MORPH                     & 6.4                                                     \\  
                                 &                                    & \textbf{Average (others)} & \textbf{7.5}                                 \\ \hline \hline
\end{tabular}
}
\end{table}



\end{document}